\documentclass[conference]{IEEEtran}
\IEEEoverridecommandlockouts
\usepackage{cite}
\usepackage{amsmath,amssymb,amsfonts}
\usepackage{algorithmic}
\usepackage{graphicx}
\usepackage{textcomp}
\usepackage{xcolor}
\usepackage{enumerate}

\def\BibTeX{{\rm B\kern-.05em{\sc i\kern-.025em b}\kern-.08em
    T\kern-.1667em\lower.7ex\hbox{E}\kern-.125emX}}
\begin{document}


\title{Classifying Directional Trajectories Near Criticality in the Three-State Majority-Vote Model with Deep Belief Networks and Bidirectional GRUs\\

}



\author{\IEEEauthorblockN{1\textsuperscript{st} Mauricio A. Valle}
\IEEEauthorblockA{\textit{Faculty of Engineering and Sciences} \\
\textit{Universidad Adolfo Ib\'{a}\~{n}ez}\\
\textit{Millennium Nucleus for Social Data Science (SODAS)}\\
Santiago, Chile \\
mauricio.valle.b@uai.cl}
\and
\IEEEauthorblockN{2\textsuperscript{nd} Gonzalo A. Ruz}
\IEEEauthorblockA{\textit{Faculty of Engineering and Sciences} \\
\textit{Universidad Adolfo Ib\'{a}\~{n}ez}\\
\textit{Millennium Nucleus for Social Data Science (SODAS)}\\
\textit{Millennium Nucleus in Data Science for}\\
\textit{Plant Resilience (PhytoLearning)}\\
Santiago, Chile \\
gonzalo.ruz@uai.cl}
}

\maketitle

\begin{abstract}
In this work, we investigate whether the latent representations learned by a Deep Belief Network (DBN) and a Bidirectional Gated Recurrent Unit (Bi-GRU) can discriminate among four dynamically distinct trajectory types in the three-state majority vote model (MV3): approach from disorder, approach from order, departure to disorder, and departure to order. The DBN, pre-trained in an unsupervised manner on static equilibrium samples via a Gaussian-Bernoulli Restricted Boltzmann Machine input layer and architecture $784 \to 4096 \to 225 \to 81$, encodes each lattice snapshot into an 81-dimensional latent vector. A t-SNE analysis of the DBN latent space reveals only partial separation of the four trajectory types, reflecting the fact that a model trained on static configurations cannot fully resolve directional temporal structure. A two-layer Bi-GRU classifier, trained on sequences of DBN-encoded snapshots of length $T = 50$, achieves near-perfect separation of all four trajectory types in its hidden state space, as confirmed by t-SNE visualization on both training and test sets. Furthermore, a sliding-window application of the trained Bi-GRU to continuous MV3 dynamics demonstrates its ability to sense the system's current dynamical regime in real-time. These results establish a principled hierarchical architecture for detecting and classifying critical transitions in agent-based opinion dynamics models.
\end{abstract}


\section{Introduction}
Agent-based models of opinion formation, when studied through the framework of statistical mechanics, exhibit phase transitions, bifurcations between ordered and disordered collective states, governed by a noise parameter that plays the role of social temperature \cite{deoliveira1992, lima2012}. These transitions are of intrinsic scientific interest because they represent tipping points: qualitative changes in collective behavior that, once crossed, are difficult to reverse. In social contexts, analogs of such tipping points include the sudden fragmentation of a previously moderate public discourse into two irreconcilable poles, the emergence of group consensus in initially heterogeneous populations, or the destabilization of a previously stable social equilibrium. Understanding the dynamical signatures that precede these transitions, and developing computational tools capable of detecting them from system observables, is therefore not only a challenge in nonequilibrium statistical physics, but a problem with direct relevance to the modeling of social systems and financial markets, where sequence-based deep models have recently been used to flag impending instabilities \cite{valle2025}.

The three-state majority vote model (MV3) provides a minimal but non-trivial arena for studying this problem. In the MV3 model, each agent on a lattice may hold one of three discrete opinions, adopting the majority opinion of its neighbors with probability $p=1 - q$ and dissenting with noise probability $q$ \cite{zubillaga2022,luz2007}. On infinite lattices, the three-state version is known to exhibit a discontinuous order-disorder transition; however, on finite lattices, such as the $28 \times 28$ system studied here, the sharp discontinuity is rounded by finite-size effects into a continuous crossover regime, producing a rich, extended critical region. This finite-size rounding is not a limitation but an opportunity: the system spends an extended portion of its trajectory in the vicinity of the critical point, generating the kind of slowly evolving, correlated dynamics that a temporal deep learning architecture can learn to recognize and classify.

The application of machine learning to phase transitions has grown substantially over the past decade, beginning with the landmark demonstration that neural networks can identify phases of matter from raw spin configurations without prior knowledge of the order parameter \cite{carrasquilla2017}. Subsequent work showed that even unsupervised methods, specifically the ``learning by confusion" scheme, can locate critical points without labeled data \cite{van2017}, and that transitions can be learned directly from dynamical rather than static data \cite{van2018}. Most of these approaches, however, analyze configurations or signals at fixed parameter values rather than full temporal trajectories evolving through a phase transition. The dynamically richer problem of anticipating a transition before it occurs, the core challenge in early warning systems, requires methods that can extract temporal structure from sequential data. Critical slowing down near tipping points produces measurable statistical signatures in time series \cite{scheffer2009}, and recurrent deep learning architectures can detect these signatures earlier and more reliably than classical indicators in ecological and climate systems \cite{bury2021}. Extending these ideas to agent-based social models is a natural but non-trivial step, requiring careful design of both the representation learning pipeline and the temporal classification architecture.

More recently, deep learning has been applied directly to opinion dynamics models. Supervised dense networks, PCA, and variational autoencoders have been employed to characterize the continuous phase transition of the majority vote model, showing that the VAE reconstruction loss behaves as an order parameter and that the correlation between real and reconstructed configurations is universal at the critical point \cite{neto2025}. Analogous methods applied to the Biswas–Chatterjee–Sen model confirm that unsupervised techniques can identify critical points on both square and triangular lattices without explicit knowledge of the transition \cite{neto2025b}. These results demonstrate that deep generative models can learn the statistical structure of opinion configurations in a phase-sensitive way. However, these approaches treat system states as static snapshots, and even dynamics-based methods stop short of resolving the \emph{direction} of a trajectory relative to the critical point, whether a system is approaching or departing criticality, and from which phase, which is precisely the information needed to anticipate a transition.

In this work, we address this gap, using the MV3 model on a two-dimensional square lattice with periodic boundary conditions as our test system. We construct quasi-static trajectory ensembles via Monte Carlo simulation with asynchronous update rules, covering four dynamically distinct trajectory types defined by their direction and origin relative to the critical point: approach from disorder, approach from order, departure toward disorder, and departure toward order. Each trajectory constitutes a temporal sequence of lattice configurations evolving slowly through parameter space, designed to isolate the dynamical signatures of critical slowing down and symmetry breaking. We then develop a two-stage deep learning pipeline in which a Deep Belief Network (DBN) built on a Gaussian-Bernoulli Restricted Boltzmann Machine (GB-RBM) input layer first encodes high-dimensional lattice snapshots into compact latent representations, and a Bidirectional Gated Recurrent Unit (Bi-GRU) network subsequently processes these latent sequences to classify the four trajectory types. Our contributions are threefold: (i) a directional trajectory ensemble that isolates the four combinations of approach and departure from both stable phases; (ii) a hierarchical DBN--Bi-GRU pipeline that separates these modes through analysis of its latent representations via t-SNE, with the DBN providing partial separation and the Bi-GRU hidden state achieving a substantially cleaner decomposition that reflects both phase and direction; and (iii) a sliding-window experiment in which the trained model tracks regime changes in real time along continuous MV3 trajectories. Together, these results indicate that combining unsupervised representation learning with recurrent sequence modeling constitutes a principled framework for characterizing criticality dynamics in opinion systems, and a concrete step toward early warning systems for critical transitions.

\section{Proposed Method and Experiments}
\begin{figure}[t]
\centerline{\includegraphics[scale=0.35]{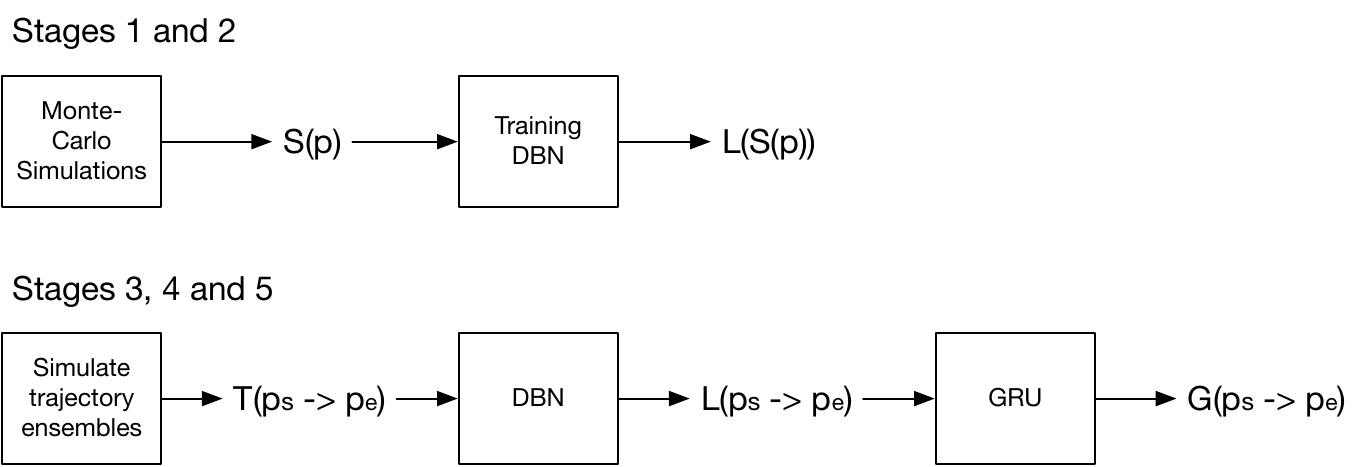}}
\caption{Methodology developed to analyze the criticality dynamics of MV3 systems.}
\label{fig1}
\end{figure}

To study the dynamics of a system governed by the majority rule, we developed a five-step methodology consisting of five stages illustrated in Figure \ref{fig1}:
\begin{enumerate}
    \item Simulating MV3 systems using Monte Carlo steps (MCS) on a lattice with periodic boundary conditions to obtain static equilibrium samples of the system at a given value of $p$, $S(p)$.
    \item Training a DBN to generate latent representations in the last hidden layer $L(S(p))$ of the samples simulated in the previous stage.
    \item Simulating ensembles of quasi-static trajectories to capture the system’s criticality dynamics. We use the notation $T(p_s \rightarrow p_e) = \{S(p_t)\}$ to denote a trajectory consisting of a sequence of system states $S(p_t)$ with different values of $p_t$ from an initial value $p_i$ to an end value $p_e$.
    \item Encode the trajectories $T(p_s \rightarrow p_e)$ using the DBN, which we will call $L(p_s \rightarrow p_e)$.
    \item Given the trajectory assembly $L(p_s \rightarrow p_e)$, train a GRU to encode its temporal structure in the embedded space of the last layer through multiclass supervised learning. We will refer to these mappings in the GRU's latent space as $G(p_s \rightarrow p_e)$.
\end{enumerate}

\subsection{MV3 System and Generation of Training Dataset} \label{sec:mv3}
We simulated the three-state majority vote model (MV3) on a
two-dimensional square lattice of $N = 28^2 = 784$ agents with
periodic boundary conditions, where each agent $i$ holds one of
three discrete opinion states $s_i \in \{0, 1, 2\}$. The energy
of the system is defined as
\begin{equation}
    H(S) = \sum_{(i,j)} \delta(s_i, s_j),
    \label{eq:energy}
\end{equation}
where $\delta$ is the Kronecker delta. At each Monte Carlo step
(MCS), agents are updated asynchronously: if the agent's four
nearest neighbors hold a majority opinion, the agent adopts it
with probability $p = 1 - q$; otherwise it dissents and assumes a
random state with probability $q = 1 - p$. If no majority exists
among neighbors, the agent adopts either competing state with
equal probability. One MCS corresponds to $N = 784$ individual
update attempts.

\begin{figure*}[t]
\centerline{\includegraphics[scale=0.63]{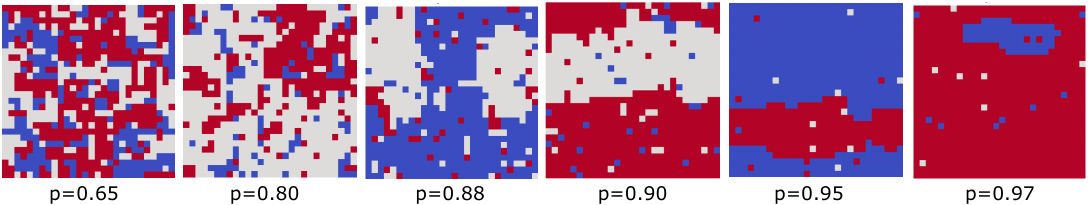}}
\caption{Examples of systems $S(p)$ of $L^2 = 28^2 = 784$ for different values of the parameter $p$, from more to less noise.}
\label{fig2}
\end{figure*}

For the static training dataset used to train the DBN, each
simulation is initialized from a random configuration and evolved
freely for at least 100,000~MCS to reach a steady state, after
which a single snapshot $S(p)$ is recorded. This process is
repeated across a broad range of values of $p$, covering
subcritical ($p < 0.75$), critical ($0.75 \leq p \leq 0.85$), and
supercritical ($p > 0.85$) regimes. The theoretical critical value is $q_c \approx 0.151$ (equivalently $p_c \approx 0.849$) \cite{lima2012},
but due to finite-size effects on the $28 \times 28$ lattice, the observed critical noise value for this system is $q_c = 0.106$ (or $p = 0.894$) at which a phase transition
is observed, characterized by the coexistence of clusters of
two or three opinion states simultaneously, giving rise to a
variety of polarization possibilities. 
Figure \ref{fig2} shows an example of $S(p)$  systems for different values of the parameter p. To calculate the system's magnetization, we use vector sum in the complex plane (similar to a Potts model with $q=3$) rather than a simple scalar. Thus, each state is represented as a 2D vector. State $s_0=0$ corresponds to a vector pointing to the right (0°), state $s_1=1$ to a vector pointing top-left (120°), and state $s_2=2$ to a vector pointing bottom-left (240°). The magnetization is defined as:


\begin{equation}
M = \frac{1}{N} \left| \sum_{i=1}^{N} e^{i\theta_i} \right| 
  = \frac{1}{N} \left| \sum_{i=1}^{N} \exp\left(i\frac{2\pi}{3}s_i\right) \right|
\end{equation}

\noindent where $N=784$, number of agents in the lattice. 
The numerator indicates  the magnitude (Euclidean length) of the net magnetization vector. Thus, when $M \approx  0$, this indicates a system in which there is complete disagreement among the agents, whereas in the opposite case, when $M \approx  1$, this indicates general agreement on one of the three states.

\subsection{Asynchronous Trajectory Ensemble Generation} \label{sec:trays}
We construct a quasi-static physics-based trajectory ensemble designed to expose the learning pipeline to the full diversity of dynamical regimes associated with the MV3 phase transition. The central motivation for a
trajectory-based approach, rather than static snapshot analysis,
is that the most diagnostically relevant information about a
system's proximity to a critical point is carried by its temporal
evolution, not by any single configuration. Near the critical
noise $q_c$, the relaxation time $\tau$ of the system diverges as
$\tau \propto |q - q_c|^{-\nu z}$, where $\nu$ is the
correlation-length critical exponent and $z$ is the dynamic
critical exponent \cite{scheffer2009, bury2021}. This phenomenon,
known as critical slowing down, means that a system traversing the
transition at any finite rate will necessarily pass through a
non-equilibrium regime near $p_c$ in which its configurations
carry memory of the preceding phase. Our ensemble is designed to systematically sample these non-equilibrium passages, generating trajectories that
encode the directional and temporal signatures of the transition
that a static ensemble cannot provide.
Four trajectory modes are defined by their direction and endpoint
relative to the observed critical point $p_c \approx 0.894$:

\begin{itemize}
    \item \textbf{Mode~0} --- From Disorder to Criticality or approach from disorder ($p\colon 0.5 \to p_c$, magnetization increases).
    \item \textbf{Mode~1} --- From Order to Criticality or approach from order
          ($p\colon 1.0 \to p_c$, magnetization decreases).
    \item \textbf{Mode~2} --- From Criticality to Disorder or depart to disorder
          ($p\colon p_c \to 0.5$, magnetization decreases);
          initialized after thermalization at $p_c$.
    \item \textbf{Mode~3} --- From Criticality to Order or depart to order
          ($p\colon p_c \to 1.0$, magnetization increases);
          initialized after thermalization at $p_c$.
\end{itemize}

The four modes are chosen deliberately to include all four
directional combinations of approach and departure relative to the
critical point, starting from both stable phases. This design
forces the classifier to learn representations that are sensitive
to the \textit{direction} of parameter evolution and not merely to
the instantaneous value of the order parameter. Modes~0 and~3
share increasing magnetization, and Modes~1 and~2 share decreasing
magnetization; distinguishing within these pairs requires
information about the history and destination of the trajectory,
which is precisely the temporal structure the GRU is designed to
exploit.

\begin{figure*}[t]
\centerline{\includegraphics[scale=0.59]{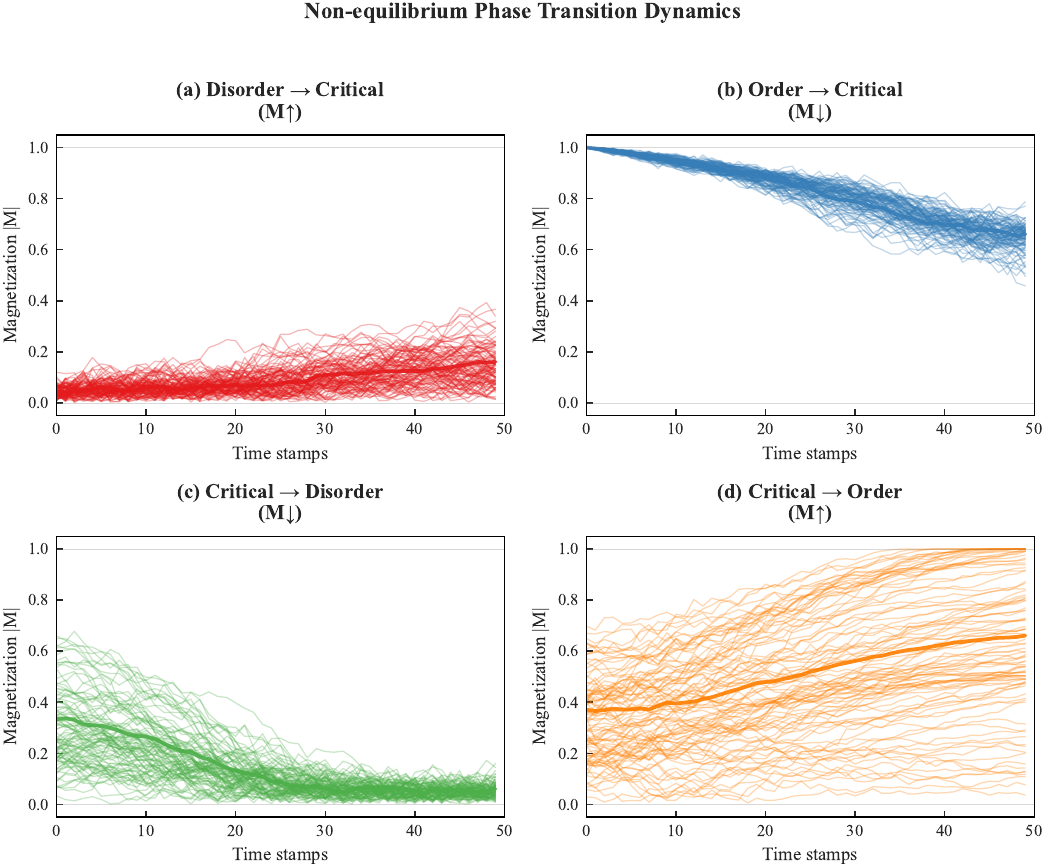}}
\caption{Example of non-equilibrium phase transition dynamics. The graph shows the magnetization evolution for 100 trajectories in each mode, with $T=50$ time steps. The thick line represents the average.}
\label{fig3}
\end{figure*}

Each trajectory $T(p_s \rightarrow p_e)$ consists of $T$ sequential snapshots
$\{S(p_t)\}_{t=1}^{T}$ for which at the start $t=1$, $p=p_s$, and at the end $t=T$, $p=p_e$. The control parameter $p$ evolves step-by-step according to a randomized phase velocity $v = \Delta p\,/\,(T \cdot u)$, where $u \sim \mathcal{U}(0.65, 0.95)$ is drawn independently for each trajectory. This \textit{asynchronous} design, inspired by the Kibble-Zurek framework for finite-rate quenches \cite{kou2023}, ensures that the critical point is crossed at a different time step in every trajectory. As a consequence, any downstream model cannot exploit a fixed temporal index as a shortcut for phase identification; it must instead detect the structural changes in the latent representations that signal the transition. Near $p_c$, the randomized velocity also naturally produces a range of non-equilibrium conditions: faster traversals generate
configurations more characteristic of the preceding phase (analogous to the impulse regime of the Kibble-Zurek mechanism), while slower traversals allow the system more time to equilibrate locally \cite{kou2023}. Modes~2 and~3, which depart from criticality, require an initial thermalization of 300~MCS at $p_c$ before the trajectory begins. Once the target value $p_{e}$ is reached, $p$ is held fixed for the remaining steps, exposing the model to examples of stationary post-transition behavior. Figure \ref{fig3} shows an example of the simulated dynamics of various possible trajectories in each of the four possible modes.

\subsubsection*{Color-Permutation Invariance and Symmetry
Augmentation.}
The MV3 model's transition rules are invariant under the
simultaneous permutation of all opinion states, a symmetry
formally described by the point group $C_{3v}$
\cite{brunstein1999, vilela2019}. This means that the three opinion
states $\{0, 1, 2\}$ are physically equivalent: no opinion label
is privileged, and any global permutation of all agents' states
produces a configuration governed by the same statistical
mechanics. In social terms, this corresponds to the fact that the
labels ``opinion A,'' ``opinion B,'' and ``neutral'' are
interchangeable, what matters is the pattern of agreement and
disagreement among agents, not the specific identity of the
majority opinion.
This symmetry has a critical practical implication for machine
learning: if the DBN and GRU are trained on trajectories in which,
say, state~$0$ is always the majority in the ordered phase, the
models may learn to recognize the specific color of the majority
rather than the structural pattern of order itself. Such a model
would fail to generalize to trajectories in which state~$1$ or
state~$2$ becomes dominant, even though these configurations are
physically identical to the training examples under a relabeling
of opinions. To enforce color-permutation invariance, each
physical trajectory is augmented with two additional permuted
copies via the mapping $s_i \mapsto (s_i + k) \bmod 3$ for
$k \in \{1, 2\}$. For every physical simulation, this produces
three latent paths through the DBN that are statistically
equivalent but span all possible majority-opinion labelings. This
augmentation triples the ensemble size without additional
simulation cost and ensures that the learned representations are
sensitive to the degree of order, the physically meaningful
quantity, rather than to which specific opinion dominates. Lattice
states are normalized to $[0, 1]$ via the mapping $\{0, 1, 2\}
\mapsto \{0.0,\,0.5,\,1.0\}$, consistent with the normalization
used during DBN training. The resulting ensemble contains
approximately 60,000 snapshots.

\subsection{Deep Belief Network and Latent Space Analysis} \label{dbn}
The DBN used in this work was pre-trained on static equilibrium
samples $S(p)$ in a previous study \cite{valle2026} and its
weights are held fixed throughout all experiments reported here.
The architecture consists of three stacked RBM layers: a
Gaussian-Bernoulli RBM (GB-RBM) as the first layer, which handles
the real-valued three-state inputs, followed by two
Bernoulli-Bernoulli RBM (BB-RBM) layers. The layer dimensions are
$784 \to 4096 \to 225 \to 81$, trained greedily layer-by-layer
using contrastive divergence \cite{hinton2002}. The GB-RBM energy
function is:
\begin{equation}
\begin{split}
    E(\mathbf{v}, \mathbf{h})  &=
        -\sum_{i,j} w_{ij} \frac{h_j}{s_j} \frac{v_i}{\sigma_i} 
        - \sum_{i}  \frac{(v_i - b_i)^2}{2\sigma_i} \\
        \qquad & - \sum_{j}  \frac{(h_j - c_j)^2}{2s_j},
    \label{eq:gbrbm}
\end{split}
\end{equation}

\noindent where $v_i$ are the static system's snapshots with values $L(S(p))$ with values $\{0.0,\,0.5,\,1.0\}$, and $h_j$ denote visible and hidden units,
respectively; $w_{ij}$ are connection weights; $b_i$ and $c_j$ are
visible and hidden biases; and $\sigma_i$, $s_j$ are standard
deviations of visible and hidden units \cite{liao2022}.
The~conditional probabilities for the GB-RBM follow these equations \cite{liao2022}:

\begin{equation}\label{eqpvh2}
    p(\mathbf{v}|\mathbf{h}) = N\Bigl( \mathbf{v}|\mathbf{b} + \sum_{j} h_j w_{ij}, \Sigma^2 \Bigr),
\end{equation}
\begin{equation}\label{eqphv2}
    p(\mathbf{h}|\mathbf{v}) = N\Bigl( \mathbf{h}|\mathbf{c} + \sum_{i} v_i w_{ij}, S^2  \Bigr)
\end{equation}

\noindent where $N(\mu, \Sigma)$ denotes the multivariate normal distribution with mean vector $\mu$ and covariance matrix $\Sigma^2$, and $S^2$ corresponds to the covariance matrix associated with the hidden units.
Both covariance matrices are kept fixed throughout training \cite{karakida}. The empirical covariance matrix computed from the training data is employed for the visible units, whereas for the hidden units we define $S = sI_m$, which implies statistical independence among hidden units. Through experimentation, we determined that $s = 0.5$ leads to the most stable convergence behavior.

Once the GBRBM has been trained, the hidden unit activations from this first layer are fed as input to train a subsequent layer using a conventional RBM. In other words, the hidden layer of the initial network acts as the visible layer for training the next RBM, which in this case is a Bernoulli-Bernoulli RBM (BBRBM). For this second model, the update rules for the conditional probabilities of both visible and hidden units follow the standard formulation \cite{fisher}. Consequently, the full DBN is constructed by stacking multiple RBMs, each trained sequentially via a greedy, layer-by-layer unsupervised learning procedure.
Parameter estimation for every RBM is carried out using contrastive divergence (CD), a widely adopted learning algorithm for RBM training \cite{hinton2002}.

\begin{figure*}[t]
\centering
\centerline{\includegraphics[scale=0.31]{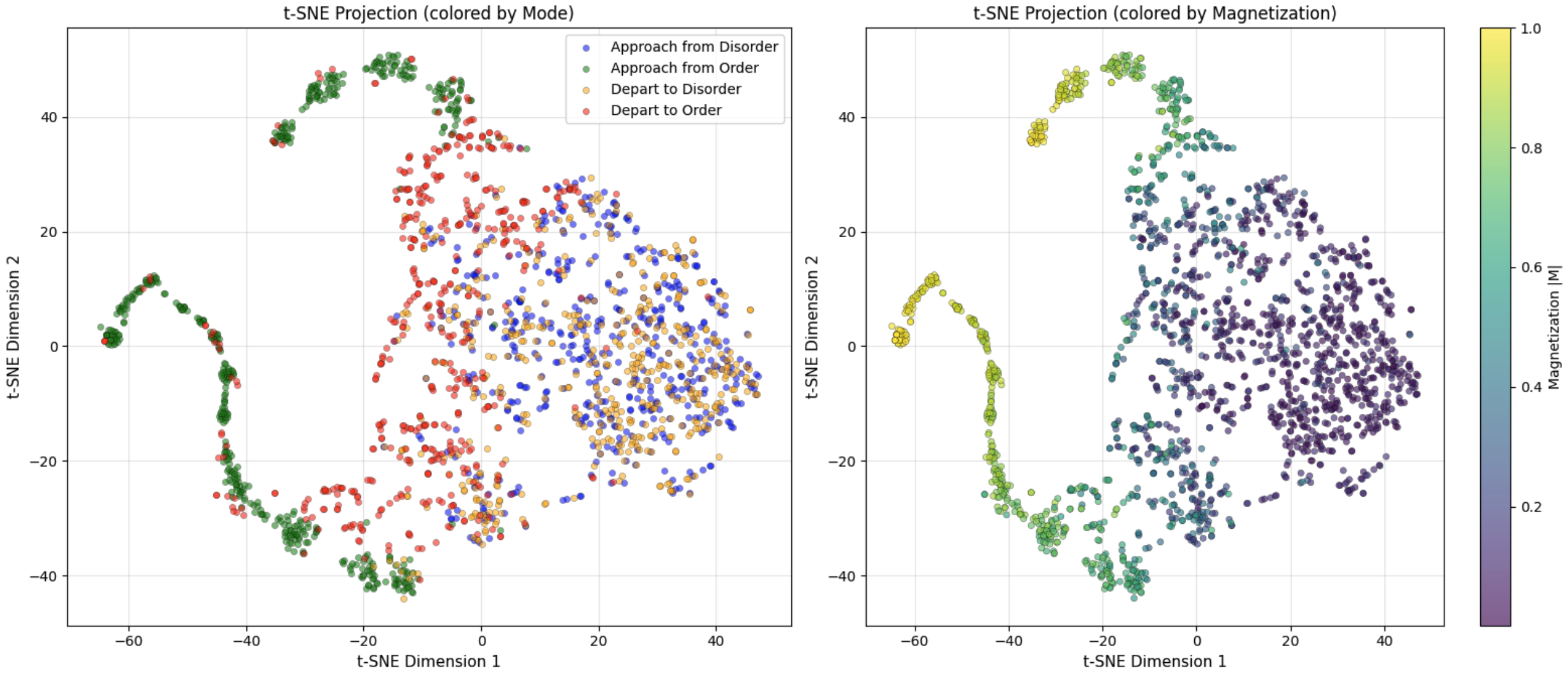}}
\caption{2D t-SNE projection of the latent representations $\mathbf{z}_t$ or $L(p_s \rightarrow p_e)$ last layer of the DBN. Left: The samples have been colored according to the trajectory mode to which they belong. Right: The samples have been colored according to the system’s magnetization value.}
\label{fig_tsn}
\end{figure*}

Once the DBN is trained, we feed trajectories $T(p_s \rightarrow p_e)$ to the first layer and let the DBN encode them by propagating them forward through all three DBN layers via the learned conditional probabilities $p(\mathbf{h}^{(l)} \mid \mathbf{h}^{(l-1)})$,
yielding an 81-dimensional latent vector $\mathbf{z}_t \in \mathbf{R}^{81}$ per snapshot. Applying a model trained on static equilibrium configurations to out-of-equilibrium trajectory samples is a deliberate design choice: it probes whether the DBN's learned representations remain informative about the dynamical regime beyond their training distribution. The discriminative power of these representations is assessed through t-SNE visualization \cite{van2008} of the full ensemble of latent vectors
$\{\mathbf{z}_t\}$ (or $L(p_s \rightarrow p_e)$) in Figure \ref{fig_tsn}, colored by trajectory mode, which reveals the degree to which the DBN latent space geometrically separates the four trajectory types prior to any temporal modeling. At this point, it is clear that the DBN alone does not achieve robust discriminative power for each of the four trajectory modes, nor is it expected to, given that the DBN was trained on static samples rather than dynamic sample sequences. To capture the dynamics, we use a Bi-GRU.

\subsection{Gated Recurrent Unit for Trajectory Classification and Dynamic Latent Representation} \label{sec:gru}
To exploit the temporal structure of the encoded trajectories, each
trajectory is reorganized as a sequential tensor of shape
$(T \times 82)$, where each row corresponds to the 81-dimensional plus one additional unit indicating the physical magnetization of the system, which represents the DBN latent representation $\mathbf{z}_t$ of one snapshot. This means that the GRU is trained to be ``physics-aware". The full ensemble is structured into a 3D input tensor of shape
$(n_{\text{traj}} \times T \times 82)$ for batch training, with
$T =50$ time steps retained in full to preserve long-range
temporal correlations that distinguish trajectory modes sharing
similar instantaneous magnetization levels.

The recurrent architecture consists of two stacked Bidirectional Gated Recurrent Unit (Bi-GRU) layers \cite{cho2014}. Processing each sequence in both forward and backward directions allows the model to integrate contextual information from the entire trajectory at each time step, which is particularly important near the critical point where the direction of approach cannot be inferred from instantaneous features alone. The first Bi-GRU layer contains 64 hidden units per direction and returns the full sequence of hidden states, while the second Bi-GRU layer contains 32 hidden units per direction. This allows the first layer to extract low-level temporal features (fluctuations) and the second to capture high-level transition signatures.
Each layer includes a dropout at 20\% regularization rate and batch normalization. Then we add a pooling layer \texttt{GlobalAveragePooling1D} which reduces the 3D tensor to a 2D output tensor calculating the average of all values of the last $T=50$ time steps before passing the output to the dense layer.

The hidden state update at each step $t$ for the forward
direction follows:
\begin{equation}
    \mathbf{h}_t = \text{GRU}(\mathbf{z}_t,\, \mathbf{h}^{g}_{t-1}),
    \label{eq:gru}
\end{equation}
\noindent where $ \mathbf{h}^{g}_{t-1}$ denotes the state of the hidden units of the Bi-GRU at the previous time step, and with the backward direction processing the sequence in reverse, and the outputs of both directions are concatenated at each layer. The final hidden state is passed to a dense layer of 32 units with \texttt{ReLU} activation, followed by a softmax output layer of 4 units, one for each trajectory mode:
\begin{equation}
    \hat{\mathbf{y}} = \text{softmax}\!\left(
        \mathbf{W}_o\,\text{ReLU}(\mathbf{W}_d\,\mathbf{h}^{g}_T +
        \mathbf{b}_d) + \mathbf{b}_o
    \right),
    \label{eq:output}
\end{equation}
where $\mathbf{W}_o$, $\mathbf{W}_d$, $\mathbf{b}_o$, and
$\mathbf{b}_d$ are learned weight matrices and bias vectors. The
model is trained as a 4-class classifier using categorical
cross-entropy loss.

\begin{figure*}[t]
\centering
\centerline{\includegraphics[scale=0.84]{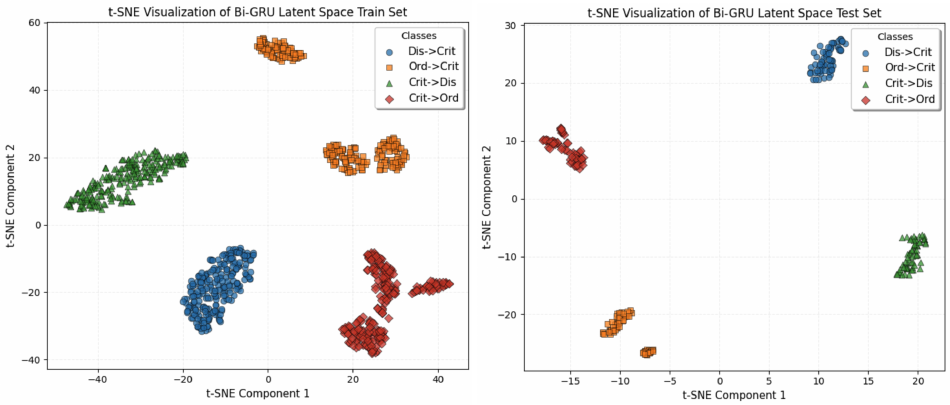}}
\caption{2D t-SNE projection of the latent representations $\mathbf{h}^{g}_t$ or $G(p_s \rightarrow p_e)$, the last layer of the 32 hidden units of the Bi-GRU, using $T=50$ time steps. Left: the resulting projection for the train set. Right: the resulting projection for the test set. These projections were performed using 960 samples from the train and test set respectively, with approximately 240 samples for each of the four trajectories.}
\label{fig_tsn_gru}
\end{figure*}

The discriminative power of the GRU's internal representations of the last 32 hidden units layer is
evaluated through t-SNE visualization \cite{van2008} of
the final hidden states $\mathbf{h}_T$ across the ensemble (equivalent to the $G(p_s \rightarrow p_e)$ in Figure \ref{fig1}),
colored by trajectory mode (See Figure \ref{fig_tsn_gru}). This analysis is performed in conjunction with the t-SNE visualization of the DBN latent vectors described in the previous Subsection, forming a paired comparison that quantifies the additional trajectory-type separation gained by processing the full temporal sequence beyond what the static DBN
encoding provides at the snapshot level.

These results may justify the encouraging performance as a classifier. The Bi-GRU classifier was trained using a dataset of 1,200 labeled synthetic trajectories. A stratified random split was employed, with 80\% of the samples allocated to the training set and the remaining 20\% reserved for testing. The Bi-GRU achieved an average classification accuracy of 1.00 on the training set and 0.99 on the test set, indicating excellent generalization performance. The training loss converged rapidly, reaching a stable minimum within approximately five epochs, with no evidence of overfitting. To assess the contribution of physically informed features, we trained an alternative Bi-GRU using the same protocol but excluding the magnetization from the DBN latent representation. Despite the removal of this information, the classifier maintained a high level of performance, achieving an average test accuracy of 0.978. Class-wise evaluation revealed recall values of at least 0.96 for all trajectory modes, while precision exceeded 0.97 in all cases except for Mode 2 trajectories (Crit $\rightarrow$ Dis), for which the precision was 0.91. These results indicate that the temporal patterns encoded in the latent representations are highly informative for trajectory classification, even in the absence of explicit magnetization information.
To assess whether the DBN outperforms the MV3 system as an input to the Bi-GRU in terms of representational capacity compared to a simpler dimension reduction technique, we applied principal component analysis (PCA) to all training samples $ \{ S(p_t) \}_{t=1}^{T}$. We then trained the Bi-GRU using the first 81 components, following the same training protocol. The average classification accuracy achieved was 0.995 and 0.854 for the training and test sets, respectively. We observed that on the test set, the precision is 0.722 for Mode 2 trajectories (Crit $\rightarrow$ Dis). 
While these results are not bad, using PCA sacrifices generalization in trajectory mode discrimination compared to our original DBN-based approach, which is vital for using the Bi-GRU to detect the transitions in which the system finds itself.

\subsection{Application of the Bi-GRU to ``sense" the system's state} \label{sec:app}
Since the Bi-GRU achieves excellent discrimination between the different types of trajectories that describe various scenarios in which the system may be entering or exiting a state of criticality, it could be useful to leverage this representational capability to “sense” the dynamics of a system that is not necessarily in equilibrium, and continuously feed sequences or ``snapshots" of these dynamics into the model so that the recurrent network can indicate the system’s current state. 

For this experiment, we artificially generated long chains of MV3 using Monte Carlo simulation with a predetermined $p$ schedule. This allows us to evaluate the model’s ability to detect phase changes in the system.
To illustrate the experiment, consider the following example: We generated a long trajectory of 500 steps for an MV3 system. The system starts in equilibrium at the critical point ($p=0.894$), and we proceed in five distinct phases to change the value of $p$ in order to produce a gradual increase in magnetization, followed by an abrupt decrease as the system enters the supercritical state. The procedure is summarized as follows:
\begin{itemize}
    \item \textbf{Phase 1: Criticality}. From $t=0$ to $t=100$, $p=0.85$ constant.
    \item \textbf{Phase 2: Ramp up}. From $t=100$ to $t=200$, increasing $p$ from 0.85 to 0.92.
    \item \textbf{Phase 3: Supercriticality}. From $t=200$ to $t=300$, $p=0.93$ constant.
    \item \textbf{Phase 4: Quench}. From $t=300$ to $t=350$, decreasing $p$ from 0.92 to 0.70. 
    \item \textbf{Phase 5: Disorder or subcriticality}: from $t=350$ to $t=500$, $p=0.70$ constant.
\end{itemize}

\begin{figure}[t]
\centerline{\includegraphics[scale=0.65]{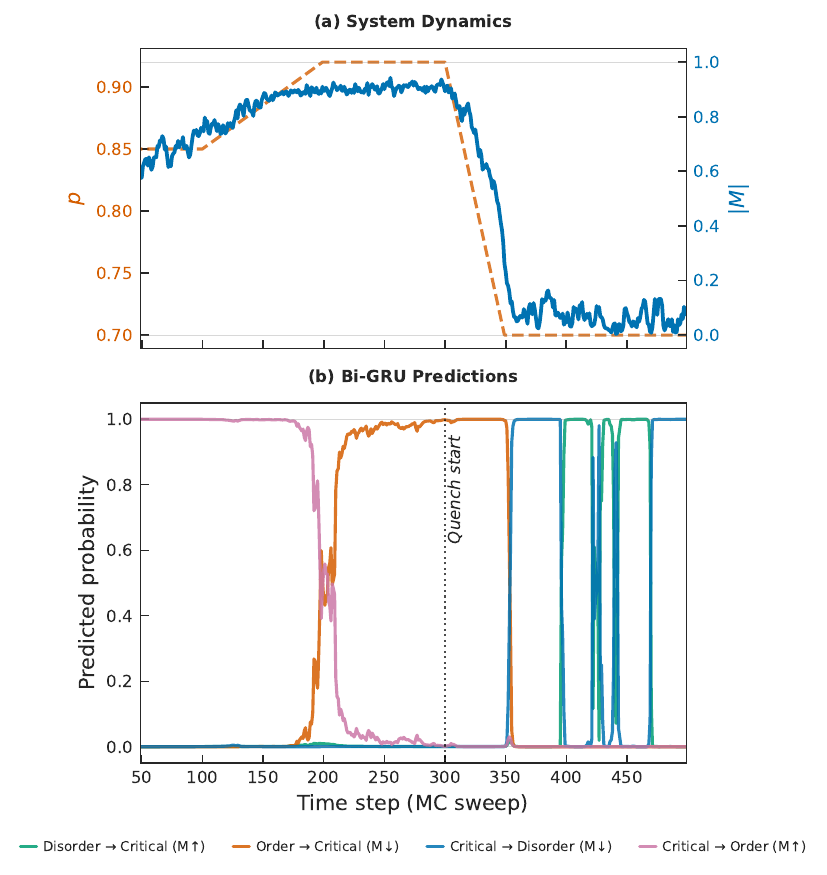}}
\caption{System dynamics and GRU predictions. (a) Evolution of the parameter $p$ (orange dashed line) and the magnetization $|M|$ (blue line). (b) Probabilities predicted by the Bi-GRU for the four phase transitions. The dotted vertical line indicates the onset of the quench.}
\label{fig_grusensing}
\end{figure}

Figure \ref{fig_grusensing}(a) shows the dynamics of the resulting  system's magnetization and the corresponding value of $p$ used to create this particular simulation. We deliberately created an abrupt transition from a supercritical state (high magnetization) to a frozen state (low magnetization) to evaluate the Bi-GRU's ability to recognize this sudden change.
Once the system dynamics have been generated, each snapshot $S(p,t)$ is fed into the DBN to obtain the corresponding latent representation in the final layer $L(S(p,t))$, yielding a sequence $\{L(S(p,t))\}_{t=1}^{500}$ of the system dynamics. From this sequence, we create sliding windows of $T=50$ time steps with 1-step overlaps (Stride = 1), which are fed into the Bi-GRU.  In this way, we allow the recurrent network to gradually sense, step by step, the type of trajectory (one of the four) on which the system is currently located. The output of the Bi-GRU is a probability vector generated by the softmax activation function, indicating the probability of each of the four trajectories.

Figure \ref{fig_grusensing}(b) shows the results of the multiclass classification. The plot confirms that the model acts as a highly sensitive phase detector rather than an anticipatory one. It ``senses" the change in real-time by identifying the transient physics currently held within the $T=50$ window.
In Phase 1 ($t < 200$), even though $p=0.85$ (criticality), the slight upward drift in magnetization $|M|$ is being picked up by the Bi-GRU as Mode 3 (Crit $\to$ Ord). This shows the GRU is sensitive to the direction of the magnetization flow, not just the static value of $p$. In the Ordered Phase Paradox ($t \approx 200-300$) is the most interesting part of the plot. As $|M|$ stabilizes at a high value, the Bi-GRU shifts to Mode 1 (Ord $\to$ Crit). This is actually a very logical ``deduction" by the model: in the training ensemble, a stable high-magnetization state is the prerequisite for an ``Order $\to$ Critical" transition. Since the system isn't increasing in order anymore, the Bi-GRU essentially ``waits" for the decay. In the Quench Detection ($t > 300$), the transition to Mode 2 (Crit $\to$ Dis) is remarkably sharp. The model doesn't predict the quench; it detects the entry of disordered frames into the sliding window. Because the Bi-GRU was trained with the magnetization as a 82nd feature, it likely triggers this class the moment the vector sum $|M|$ starts to tilt back toward the origin.

\section{Conclusions}

This work investigated the discriminative power of a hierarchical deep learning pipeline, combining an unsupervised Deep Belief Network encoder with a supervised
Bidirectional Gated Recurrent Unit classifier, for identifying four dynamically distinct trajectory types in the three-state majority vote model. The central finding
is that neither static latent representations nor temporal sequence modeling alone is sufficient: the DBN, applied out of distribution to non-equilibrium trajectory snapshots, produces only partial separation of the four trajectory modes in t-SNE space, with modes sharing the same direction of magnetization change remaining difficult to distinguish.
The Bi-GRU, by contrast, achieves near-perfect separation of all four modes in its hidden state space, demonstrating that the temporal structure of the DBN-encoded sequences carries the directional and historical information that instantaneous configurations cannot provide. The sliding-window sensing experiment further shows that the trained Bi-GRU can track the system's dynamical regime in real-time along continuous MV3 trajectories, correctly identifying transitions between phases and responding sharply to abrupt quenches. Together, these results confirm
that the combination of unsupervised representation learning on equilibrium configurations with recurrent sequence modeling of non-equilibrium trajectories constitutes an effective and physically principled framework for characterizing criticality dynamics in opinion systems.

Some methodological limitations must be acknowledged. First, the DBN was trained exclusively on static equilibrium samples, and its application to out-of-equilibrium
trajectory snapshots introduces an intrinsic out-of-distribution risk: the learned representations may not capture all dynamically relevant features of non-equilibrium configurations, particularly near the critical point where the system departs most strongly from its training distribution. 
Second, all experiments are conducted on a finite $28 \times 28$ lattice, whose finite-size effects round the discontinuous bulk transition into a continuous crossover and shift the effective critical point. The degree to which the pipeline's discriminative power generalizes to larger lattices, other network topologies, or systems with different universality classes remains an open question. 

Future work will pursue two primary directions. The first is to reframe the Bi-GRU training objective from classification of the current trajectory mode, a
\textit{sensing} task, toward anticipation of an impending transition, that is, an \textit{early warning} task in which the model must assign elevated probability to an approaching critical crossing before it is observed in the sliding window \cite{valle2025}. This requires redesigning the training labels to reflect proximity to the critical point rather than trajectory identity, and may benefit from incorporating uncertainty quantification to provide calibrated warning signals. The second direction is to apply the full pipeline to real-world time series from social and financial systems, such as sequences of collective sentiment indicators, social network activity patterns, or financial return volatility, where the existence and timing of critical transitions is not known a priori. Such an application would test the extent to which representations learned from a stylized model can transfer to the richer, noisier dynamics of empirical data, and would constitute a concrete step toward machine learning--driven early warning systems for collective social behavior.

\section*{Acknowledgment}
The authors thank ANID FONDECYT 1250386, ANID FONDECYT 1230315, ANID-MILENIO-NCN2024\_103, ANID-MILENIO-NCN2024\_047, and Centro de Modelamiento Matemático (CMM) FB210005, BASAL funds for centers of excellence from ANID-Chile.

\bibliographystyle{IEEEtran}
\bibliography{example}

\end{document}